\documentclass[runningheads]{llncs}
\usepackage{graphicx}
\usepackage{hyperref}
\usepackage{amsmath}
\usepackage{amssymb}
\usepackage{booktabs}
\usepackage{subcaption}
\usepackage{url}
\usepackage{marvosym}

\begin{document}
\title{Multi-Label Out-of-Distribution Detection with Spectral Normalized Joint Energy}
\titlerunning{Spectral Normalized Joint Energy}
% \author{Anonymous Authors}
% \institute{Double-Blind Review}
\author{Yihan Mei\inst{1}\orcidID{0009-0001-6050-2147} \and
Xinyu Wang\inst{1}\orcidID{0009-0005-1948-5065} \and
Dell Zhang\inst{2}\orcidID{0000-0002-8774-3725} \and
Xiaoling Wang\inst{1}\textsuperscript{(\Letter)}\orcidID{0000-0002-4594-6946}}
\authorrunning{Y. Mei et al.}
\institute{
    East China Normal University, Shanghai 200062, China\\
    \email{\{yihanmei,xinyu\_wang\}@stu.ecnu.edu.cn}\\
    \email{xlwang@cs.ecnu.edu.cn} \and
    Institute of Artificial Intelligence (TeleAI), China Telecom Corp Ltd, China\\
    \email{dell.z@ieee.org}
    }
\maketitle              % typeset the header of the contribution
\begin{abstract}

In today's interconnected world, achieving reliable out-of-distribution (OOD) detection poses a significant challenge for machine learning models. 
While numerous studies have introduced improved approaches for multi-class OOD detection tasks, the investigation into \emph{multi-label} OOD detection tasks has been notably limited. 
We introduce Spectral Normalized Joint Energy (SNoJoE), a method that consolidates label-specific information across multiple labels through the theoretically justified concept of an energy-based function. 
Throughout the training process, we employ \emph{spectral normalization} to manage the model's feature space, thereby enhancing model efficacy and generalization, in addition to bolstering robustness. 
Our findings indicate that the application of spectral normalization to \emph{joint energy} scores notably amplifies the model's capability for OOD detection. 
We perform OOD detection experiments utilizing PASCAL-VOC as the in-distribution dataset and ImageNet-22K or Texture as the out-of-distribution datasets. 
Our experimental results reveal that, in comparison to prior top performances, SNoJoE achieves 11\% and 54\% relative reductions in FPR95 on the respective OOD datasets, thereby defining the new \emph{state of the art} in this field of study.

\end{abstract}

\section{Introduction}

In the current digital era, the pervasive use of machine learning models is undeniable. 
However, these models often grapple with data that deviates from their training data, known as out-of-distribution (OOD) data, when deployed in real-world settings. 
This discrepancy can lead to inaccurate predictions, raising safety concerns and other issues. 
OOD detection plays a crucial role in identifying unfamiliar data, thereby enhancing model safety and robustness in diverse environments. 
Thus, assessing OOD uncertainty emerges as a critical challenge for researchers.

Significant advancements have been made in OOD detection research. 
The Local Outlier Factor (LOF) method~\cite{10.1145/335191.335388} and unsupervised outlier detection using globally optimal sample-based Gaussian Mixture Models (GMM) by Yang et al.~\cite{yang2009outlier} represent foundational work. 
G-ODIN~\cite{hsu2020generalized} builds on ODIN~\cite{liang2020enhancing} to improve sensitivity to covariate shifts. 
OpenMax~\cite{bendale2015open} introduces Extreme Value Theory (EVT) to neural networks, calibrating logits with EVT probability models, including the Weibull distribution. Classification-based approaches see innovations like extending One-Class Classification (OCC) through elastic-net regularization for learning decision boundaries~\cite{reiss2021panda}, and selecting reliable data from unlabeled sources as negative samples for supervised anomaly detection settings~\cite{chaudhari2012learning}.

Despite these advancements, OOD detection in multi-label classification contexts remains underexplored. 
Multi-label classification poses unique challenges due to the necessity of evaluating uncertainty across multiple labels, rather than a single dominant one~\cite{wang2021multilabel}. Achieving stable model training is essential for accurate multi-label OOD sample identification, with strategies like using free energy for OOD uncertainty assessment proposed by Liu et al.~\cite{liu2021energybased}.

This paper introduces a novel approach, \textbf{S}pectral \textbf{No}rmalized \textbf{Jo}int \textbf{E}nergy (SNoJoE), for assessing OOD uncertainty in multi-label datasets. 
SNoJoE calculates free energy for each label and combines these energies, overcoming the difficulties generative models face in estimating joint likelihood for multi-label data~\cite{hinz2019generating}. 
Additionally, it demonstrates that aggregating label energies is more effective than summing label scores in OOD detection evaluations~\cite{wang2021multilabel}, highlighting the importance of choosing the right label assessment function.

We also utilize ResNet for feature extraction from in-distribution images, employing an energy function as the metric for OOD assessment. 
To counter overfitting and enhance model robustness, we apply spectral normalization as a regularization technique. 
Our findings show that spectral normalization reduces gradient variation ranges during training, minimizing the risk of gradient problems and promoting a well-regulated feature space. 
This approach helps the model to generalize better to OOD instances by focusing on extracting generalizable features rather than memorizing training data. 
Applying spectral normalization to OOD detection tasks has been shown to significantly improve model performance, such as achieving a significant 54\% reduction in FPR95 on the Texture dataset with respect to PASCAL-VOC ($t$-test $p\text{-value} < 0.01$), underscoring the technique's value in OOD detection. 

Our main contributions include:
\begin{itemize}
    \item Introducing SNoJoE, an innovative method for OOD uncertainty assessment in multi-label classification that can deliver today's best performance on two real-world datasets.
    \item Demonstrating through ablation studies that spectral normalization significantly enhances multi-label OOD detection performance.
    \item Making our experimental code and datasets available for reproducible research\footnote{\url{https://github.com/Nicholas-Mei/Ood_Detection_SNoJoE}}.
\end{itemize}

 % reducing the average FPR95 by 4.71\% and improving AUROC and AUPR by 1.43\% and 1.18\%, respectively.

\section{Related Work}

\subsection{Multi-Label Classification}

Unlike the simpler scenario of multi-class (single-label) classification, multi-label classification allows each image to be associated with multiple label concepts. 
Early approaches to multi-label classification treated the presence of each label independently, neglecting the potential correlations among labels~\cite{gong2014deep,7305792}.

Initial research in multi-label classification demanded significant computational resources. 
Ghamrawi and McCallum~\cite{ghamrawi2005collective} employed Conditional Random Fields (CRF) to create graphical models that identify correlations between labels, and Chen et al.~\cite{chen2015learning} integrated CRF with deep learning techniques to examine the dependencies among output variables. 
These strategies necessitate the explicit modeling of label correlations, leading to elevated computational demands.

Conversely, deep learning techniques do not inherently require substantial computational resources for multi-class recognition tasks and have shown notable effects~\cite{wang2016cnnrnn,tsoumakas2007multi}. 
Gong et al.~\cite{gong2014deep} utilized Convolutional Neural Networks (CNN) to label images with 3 or 5 labels in the NUS-WIDE dataset, while Chen et al.~\cite{8708965} applied CNNs to categorize road scene images from a set of 52 potential labels. 
Thus, efficiently solving multi-label classification challenges is intricately linked to a wide range of applications in the contemporary open world.

\subsection{Out-of-Distribution Detection}

In the realm of OOD detection, research has primarily concentrated on four areas: Novelty Detection (ND), Open Set Recognition (OSR), Outlier Detection (OD), and Anomaly Detection (AD).

Initially, methods leaned heavily on confidence estimation and the setting of thresholds, judging inputs' relevance to known categories by the confidence scores produced by the model. 
However, they often falter when facing complex data distributions. 
Zhang et al.~\cite{zhang2020hybrid} introduced OpenHybrid, a strategy that combines representation space learning from both an inlier classifier and a density estimator, the latter acting as an outlier detector.

Ayadi et al.~\cite{ayadi2017outlier} outlined twelve diverse interpretations of outliers, highlighting the challenge of defining outliers precisely. 
This has spurred a wave of innovative approaches for identifying and addressing outliers~\cite{ranshous2015anomaly}. 
Among them, density-based methods for detecting outliers represent some of the earliest strategies. 
The Local Outlier Factor (LOF) method, introduced by Breunig et al.~\cite{10.1145/335191.335388}, stands as a pioneering density-based clustering technique for outlier detection, leveraging the concept of loose correlation through k-nearest neighbors (KNN). 
LOF calculates local reachability density (LRD) within each point's KNN set and compares it to the densities of neighbors within that set.

Yang et al.~\cite{yang2009outlier} proposed an unsupervised outlier detection approach using a globally optimal sample-based Gaussian Mixture Model (GMM), employing the Expectation-Maximization (EM) algorithm for optimal fitting to the dataset. 
They define an outlier factor for each data point as the weighted sum of mixture proportions, where weights denote the relationships among data points.

Certain studies have focused on increasing sensitivity to covariate shifts by examining hidden representations in neural networks' intermediate layers. 
Generalized ODIN~\cite{hsu2020generalized} builds on ODIN~\cite{liang2020enhancing} by adopting a specialized training objective, DeConf-C, and choosing hyperparameters like perturbation magnitude for in-distribution data. 
Wei et al.~\cite{wei2022mitigating} demonstrated that issues of overconfidence could be alleviated through Logit Normalization (Logit Norm), which counters the typical cross-entropy loss by enforcing a constant vector norm on logits during training, enabling neural networks to distinctly differentiate between in-distribution and OOD data. 
Other efforts have sought to refine OOD uncertainty estimation via confidence scores based on Mahalanobis distance~\cite{lee2018simple} and gradient-based GradNorm scores~\cite{huang2021importance}.

Within classification-based OOD detection methods, One-Class Classification (OCC) uniquely establishes a decision boundary matching the expected normal data distribution density level set~\cite{tax2002one}. 
Deep SVDD~\cite{pmlr-v80-ruff18a} was the first to adapt classical OCC for deep learning, mapping normal samples to a hypersphere to delineate normality. 
Deviations from this model are flagged as anomalous. Later efforts expanded this approach through elastic regularization~\cite{reiss2021panda} or adaptive descriptions with multi-linear hyperplanes~\cite{wang2019gods}. 
Additionally, some methods employ Positive-Unlabeled (PU) learning in semi-supervised AD contexts, providing unlabeled data alongside normal data. 
Mainstream PU strategies either select reliable negative samples for a supervised AD setting, using clustering~\cite{chaudhari2012learning} and density models~\cite{he2020instancedependent}, or treat all unlabeled data as noise negatives for learning with noise labels, employing sample re-weighting~\cite{pmlr-v37-menon15} and label cleaning~\cite{pmlr-v38-scott15,zhong2019graph}.

Despite advancements, OOD detection remains a challenging field, predominantly explored within multi-class tasks, with limited work in multi-label classification. 
Hence, we introduce a technique that integrates spectral normalization into the network and utilizes energy scores to derive label-wise joint energy scores for OOD detection tasks.

\subsection{Energy-based Models}

Energy-based models (EBMs) in machine learning trace their origins to Boltzmann machines~\cite{ackley1985learning}. 
This approach offers a cohesive framework encompassing a broad spectrum of learning algorithms, both probabilistic and deterministic~\cite{lecun2006tutorial,ranzato2006efficient}. 
Xie et al.~\cite{xie2016theory} showed that the discriminative classifiers within GAN networks can be interpreted through an energy-based lens. 
Moreover, these methods have been leveraged for structured prediction challenges~\cite{tu2018learning}.

Recent studies~\cite{liu2021energybased,lin2021mood} have advocated for the use of energy scores in detecting OOD instances, grounding their arguments in theoretical perspectives related to likelihood~\cite{morteza2021provable}. 
Here, samples exhibiting lower energy are classified as in-distribution (ID), while those with higher energy are flagged as OOD. 
Liu et al.~\cite{liu2021energybased} pioneered a technique for quantifying OOD uncertainty by utilizing energy scores, showcasing remarkable efficacy in multi-class classification networks. 
Meanwhile, research by Wang et al.~\cite{wang2021multilabel} targets multi-label contexts, illustrating the benefits of harnessing the collective power of all label data. 
Our contribution merges cross-label energy scores, affirming enhanced performance through the implementation of spectral normalization.

\section{Method}

In this section, we introduce a novel approach for OOD detection in multi-label scenarios. 
First, we address multi-label inputs by integrating concepts from the free energy function, assessing OOD uncertainty through the evaluation of joint label energies across labels. Subsequently, we present SNoJoE, a technique that applies spectral normalization to the joint label energy scores. 
This enhancement not only improves the model's robustness but also facilitates the extraction of features that are more generalizable.

\subsection{Preliminaries}

\subsubsection{Multi-label Classification}

Multi-label classification is a machine learning task where the goal is to assign input data samples to one or more categories out of a set of predefined labels. Unlike traditional single-label classification tasks, where each sample can only belong to one category, multi-label classification allows a sample to have multiple labels simultaneously.
Generally, consider $\mathcal{X}$ (representing the input space) and $\mathcal{Y}$ (representing the output space), with $\mathcal{P}$ denoting a distribution over $\mathcal{X} \times \mathcal{Y}$. Suppose $f: \mathcal{X} \longrightarrow \mathbb{R}^{\left | \mathcal{Y} \right |}$ represents a neural network trained on samples drawn from P. Each input can be correlated with a subset of labels in $\mathcal{Y} = {1, 2, \cdots, K}$, denoted by a vector $\mathrm{\textbf{y}} = [y_1, y_2, \cdots, y_K]$, where 
\begin{equation}
    y_i =
\begin{cases}
 1  \text{ , if }i\text{ is associated with } x\\
 0  \text{ , otherwise } 
\end{cases}.
\end{equation}
Utilizing a convolutional neural network (CNN) with a shared feature space, we generate multi-label output predictions. This approach has emerged as the standard training mechanism for multi-label classification tasks, finding widespread application across various domains~\cite{zhang2018deep,liu2015multi}.

\subsubsection{Out-of-distribution Detection}

Similar to the concept presented in ~\cite{wang2021multilabel}, we define the problem of OOD detection for multi-label classification as follows. Let $D_{in}$ denote the marginal distribution $\mathcal{P}$ over the label set $\mathcal{X}$, representing the distribution of in-distribution data. During testing, the environment may generate out-of-distribution data $D_{out}$ on $\mathcal{X}$ . 
The goal of OOD detection is to define a decision function $D$ such that:
\begin{equation}
D(x;f)=
\begin{cases}
 1  \text{ , if x}\sim \mathcal{D}_{in} \\
 0  \text{ , if x}\sim \mathcal{D}_{out} 
\end{cases}.
\end{equation}

\subsubsection{Energy Function}
The definition of the energy equation was first proposed by Liu et al. They introduced the free energy as the scoring function for OOD uncertainty assessment in a multi-class setting. Given a classifier $f(\mathrm{x}):\mathcal{X} \rightarrow \mathbb{R}^K$ mapping the input $x$ to $K$ real numbers as logits, the class distribution is represented through softmax:
 \begin{equation}
     p(y_i=1|x)=\frac{e^{f_{y_i}(x)}}{\Sigma ^K_{j=1}e^{f_{y_j}(x)}}.
 \end{equation}
Then, the transformation from logits to probability distribution is achieved through the Boltzmann distribution: 
\begin{equation}
    p(y_i=1|x)=\frac{e^{-E(x,y_i)}}{\int_{y^\prime}e^{-E(x,y^\prime)}}=\frac{e^{-E(x,y_i)}}{e^{-E(x)}}.
\end{equation}
Thus, the initially defined classifier can be interpreted from an energy-based perspective. Viewing the logits $f_{y_i}(x)$ as an energy function $E(x,y_i)$, we can obtain the free energy function $E(x)$ for any given input $\mathrm{x}$:
\begin{equation}
    E(x)=-\log\sum^K_{i=1}e^{f_{y_i}(x)}.
\end{equation}

\subsection{Label-wise Joint Energy}
\label{JoD}
We first consider the problem of OOD uncertainty detection on a standard multi-label classifier. For a given input $x$, its output for the $i$-th class is: 
\begin{equation}
\label{eq:f}
    f_{y_i}(x)=h_{l-1}(x)\cdot w^i_{cls}
    \textrm{,}
\end{equation}
where $h_{l-1}(x)$ is the feature vector of the penultimate layer of the network, and $w^i_{cls}$ is the weight matrix corresponding to $i$-th class. The predictive probability of label $y_i$ is then implemented through a variant of a binary logistic classifier:
\begin{equation}
\label{eq:p1}
    p(y_i=1\ |\ x)=\frac{e^{f_{y_i}(x)}}{1+e^{f_{y_i}(x)}} \textrm{.}
\end{equation}
For the logistic form in equation~\ref{eq:p1}, we can consider it as a softmax form with only $0$ and $e^{f_{y_i}(x)}$ as the logits:
\begin{equation}
\label{eq:p2}
    p(y_i=1\ |\ x)=\frac{e^{f_{y_i}(x)}}{e^0+e^{f_{y_i}(x)}}
    \textrm{.}
\end{equation}
Through the softmax form of equation~\ref{eq:p2}, for each $i \in \{1, 2, ..., K\}$, the \textit{energy function} of class $y_i$ can be expressed as follows:
\begin{equation}
\label{eq:en}
    E_{y_i}(x)=-\ln(1+e^{f_{y_i}(x)})
    \textrm{.}
\end{equation}
Therefore, for each class $\{y_i\}_{i=1}^K$, we can derive a \textit{label-wise joint energy function} as follows:
\begin{equation}
\label{eq:joint}
    E_{joint}(x)=\sum_{i=1}^{K}-E_{y_i}(x)
\end{equation}
In Equation~\eqref{eq:en}, we consider the joint uncertainty among labels. 
Wang et al.~\cite{wang2021multilabel} provided a theoretical foundation based on joint likelihood. 
Subsequent work by Zhang and Taneva-Popova~\cite{zhang2023theoretical}, however, found that while Wang et al.'s approach assumed label independence, contrary to the initial beliefs of leveraging label independencies, joint energy indeed provides the optimal probabilistic approach to address the multi-label OOD problems.
Moreover, Wang et al.~\cite{wang2021multilabel} confirmed that utilizing multiple dominant labels to signal in-distribution inputs effectively captures data features, thus bypassing the need for direct computation and optimization in multi-label datasets. 
This approach also sidesteps the complexities associated with estimating joint likelihood through generative models, a notably challenging endeavor.

After deriving the \textit{label-wise joint energy} in equation~\ref{eq:joint}, we can utilize this method to detect the OOD uncertainty:
\begin{equation}
\label{eq:detect}
    D(x;\tau)={
    \begin{cases}
        \textrm{out}\ \ \ \ \ \ \ \ \textrm{if}\ E_{joint}(x)\le\tau \\
        \textrm{in}\ \ \ \ \ \ \ \ \ \ \textrm{if}\ E_{joint}(x)>\tau
    \end{cases}
    }
\textrm{,}
\end{equation}
where $\tau$ is the energy threshold. In our experimental setup, we defined $\tau=95\%$ to ensure that $D(x;\tau)$ can correctly classify the majority of in-distribution data.

\subsection{Spectral Normalized Joint Energy}
Based on the foundation laid by Section~\ref{JoD}, we present \textbf{S}pectral \textbf{No}rmalized \textbf{Jo}int \textbf{E}nergy(SNoJoE). As part of the feature vector extraction process, spectral normalization is applied to the initial layers of the model. Through power iteration, we evaluate the spectral norm, guaranteeing that the weight matrices of the model adhere to \textit{bi-Lipschitz constraint}.

Firstly, we need to ensure that the spectral norm of the weight matrices $g_l(x)=\sigma(W_lx+b)$ in the non-linear residual blocks of the network is less than 1, thereby ensuring:
\begin{equation}
\label{eq:biLips}
    \left \| g_l \right \| _{Lipschitz} \le \left \| W_lx+b \right \|_{Lipschitz} \le \left \| W_l \right \| _2 \le 1
    \textrm{.}
\end{equation}
To achieve this, we apply \textit{spectral normalization} to constrain the weight matrices of the first $L$ layers in the network:
\begin{equation}
\label{eq:sn}
    W_l={\begin{cases}
W_l/\sigma\ \ \ \ \ \ \ 1\le l\le L, \\
W_l\ \ \ \ \ \ \ \ \ \ \ l > L
\end{cases}}
\textrm{,}
\end{equation}
where $\sigma$ is the spectral norm of the weight matrix, defined as the maximum singular value of the weight matrix. This singular value is obtained through singular value decomposition(SVD) of the weight matrix.
As recommended in ~\cite{behrmann2019invertible}, \textit{spectral normalization} is used to enforce the weight matrices $\{W_l\}_{l=1}^{L}$ in equation~\ref{eq:biLips} to be \textit{Lipschitz-constrained}, ensuring that the hidden layer parameters $h_i(x)$ ``\textit{distance preserving}''.

~\cite{bartlett2018representing} demonstrates that consider a \textit{hidden mapping} $h: \mathcal{X}\longrightarrow\mathcal{Y}$ with residual architecture $h=h_{l-1}\circ \cdots \circ h_2 \circ h_1(x)$ where $h_l(x)=x+g_l(x)$. If for $0<\alpha \le 1$, all $g_l$'s are $\alpha$-\textit{Lipschitz}, \textit{i.e}., $\left \| g_l(x)-g_l(x')\right \|_Y \le \alpha \left \| x-x'\right \| _X\ \ \ \forall(x,x')\in \mathcal{X}$. Then:
\begin{equation}
\label{eq:biLips_concept}
    Lips_{lower}\ast \left \| x-x'\right \|_X \le \left \| h(x)-h(x')\right \|_Y \le Lips_{upper} \ast \left \| x-x'\right \|_X
    \textrm{,}
\end{equation}
where $Lips_{lower}=(1-\alpha)^{L-1}$ and $Lips_{upper}=(1+\alpha)^{L-1}$ are respectively the lower and upper bounds of \textit{Lipschitz continuity}. Through the \textit{bi-Lipschitz constraint}, the upper bound prevents overfitting during model gradient updates, ensuring the generalization and robustness of the model. The lower bound ensures that there is a certain distance maintained between input feature vectors, \textit{i.e.}, $h(x)$ is \textit{distance preserving}, thereby enabling the extraction of more generalizable features.

Combining the approach from Section~\ref{JoD}, we now update the expression for $h_i(x)$ in equation~\ref{eq:f}:
\begin{equation}
\label{eq:update}
    h_i(x)={\begin{cases}
\frac{W^{i-1}}{\sigma} \cdot h_{i-1}(x)\ \ \ \ \ \ \ 2\le i\le L, \\
W_{i-1} \cdot h_{i-1}(x)\ \ \ \ \ \ \ \textrm{otherwise}
\end{cases}}
\textrm{.}
\end{equation}

Through the transformation of $h_i(x)$ in equation~\ref{eq:update}, the feature vectors can possess the property of ``\textit{distance preserving}'' and replace $h_i(x)$ in equation~\ref{eq:f} to complete the subsequent OOD uncertainty detection.

\section{Experiments}
In this section, we expound upon our experimental configuration (Section~\ref{setup}) and showcase the effectiveness of our approach across various out-of-distribution (OOD) evaluation tasks (Section~\ref{results}). Furthermore, we delve into ablation studies and conduct comparative analyses, thereby fostering a deeper comprehension of distinct methodologies and ultimately contributing to an enhanced understanding of the field.

\subsection{Setup}
\label{setup}

\subsubsection{In-Distribution Datasets}
We consider the PASCAL-VOC~\cite{everingham2015pascal} as the in distribution multi-label dataset. It comprises 22,531 images of objects from 20 different categories such as \textit{people, dogs, cars, etc.}, with detailed annotations provided. In this paper, We conduct the OOD detection task to evaluate the performance of our proposed method on this dataset.

\subsubsection{Training Details}
In this study, the multi-label classifier trained is based on the ResNet-101 backbone architecture. The classifier is pretrained on ImageNet-1K~\cite{deng2009imagenet}, and the last layer is replaced by two fully connected layers. Spectral normalization is applied to the first 9 layers of the model. We utilize the Adam optimizer with standard parameters($\beta_1=0.9,\beta_2=0.999$), and the initial learning rate during training is set to $1 \times 10^{-4}$. Data augmentation techniques such as random cropping and random flipping are employed during training to enhance the dataset, resulting in color images of size $256\times 256$. After training, the mean Average Precision (mAP) on PASCAL-VOC is 89.19\%. The entire experimental process is conducted on NVIDIA GeForce RTX 2080Ti.

\subsubsection{Out-of-Distribution Datasets}
To evaluate the performance of the model trained on the in-distribution dataset, we employ the Texture dataset~\cite{cimpoi2013describing} and designate 20 classes from ImageNet-22K as out-of-distribution (OOD) datasets. Following the evaluation protocol outlined in \cite{wang2021multilabel}, we configure the ImageNet-22K dataset in a identical manner for evaluating the PASCAL-VOC pretrained model. 
The selected classes for evaluation encompass a diverse range, 
including
% \texttt{dolphin, deer, bat, rhino, raccoon, octopus, giant clam, leech, venus flytrap, cherry tree, Japanese cherry blossoms, redwood, sunflower, croissant, stick cinnamon, cotton, rice, sugar cane, bamboo, turmeric}.
\textit{dolphin, deer, bat, rhino, raccoon, octopus, giant clam, leech, venus flytrap, cherry tree, Japanese cherry blossoms, redwood, sunflower, croissant, stick cinnamon, cotton, rice, sugar cane, bamboo, turmeric}.
% \texttt{deer}, \texttt{bat}, \texttt{rhino}, \texttt{raccoon}, \texttt{octopus}, \texttt{giant clam}, \texttt{leech}, \texttt{venus flytrap}, \texttt{cherry tree}, \texttt{Japanese cherry blossoms}, \texttt{redwood}, \texttt{sunflower}, \texttt{croissant,} \texttt{stick cinnamon}, \texttt{cotton}, \texttt{rice}, \texttt{sugar cane}, \texttt{bamboo}, and \texttt{turmeric}.

\begin{table}[!tb]
\caption{The dataset configuration in experiments.}
\label{tab:image-dataset}
\centering
\setlength{\tabcolsep}{4pt} 
\begin{tabular}{lllc}
\toprule
\textbf{Dataset} & \textbf{Role} & \#\textbf{Classes} & \#\textbf{Instances}\\ 
\midrule
PASCAL-VOC\cite{everingham2015pascal} & In-Distribution (ID)      & 20                & 22,531 \\
ImageNet-22K                          & Out-of-Distribution (OOD) & 20 (out of 21841) & 18,835 \\
Texture\cite{cimpoi2013describing}   & Out-of-Distribution (OOD) & 47                &  5,640 \\
\bottomrule
\end{tabular}
\end{table}

\subsubsection{Evaluation Metrics}
In our experiments, we employ commonly used evaluation metrics for OOD detection under multi-label settings:
(i) the false positive rate (FPR95) of OOD examples is calculated when the true positive rate (TPR) of in-distribution examples is held constant at 95\%; 
(ii) the area under the receiver operating characteristic curve (AUROC); 
(iii) the area under the precision-recall curve (AUPR).

\subsection{Results}
\label{results}

\begin{table}[!tb]
\caption{The comparison of OOD detection performance using spectral normalized joint energy vs. competitive baselines. We use ResNet~\cite{he2016identity} to train on the in-distribution dataset and use ImageNet-22K (20 classes), Texture~\cite{cimpoi2013describing} as OOD datasets. 
Besides, $\dagger$ denotes that SNoJoE is statictically better (t-test with p-value $<$ 0.01) than JointEnergy on Texture~\cite{cimpoi2013describing}. 
All values are percentages. \textbf{Bold} numbers are superior results. $\uparrow$ indicates larger values are better, and $\downarrow$ indicates smaller values are better. }
\begin{subtable}[b]{\textwidth}
\caption{$\mathcal{D}_{out}$ = ImageNet-22K}
\label{tab:image-ood}
\centering
\setlength{\tabcolsep}{8pt} 
\begin{tabular}{lccc}
\toprule
\textbf{OOD Score} & \textbf{FPR95}$\downarrow$ & \textbf{AUROC}$\uparrow$ & \textbf{AUPR}$\uparrow$ \\ 
\midrule
MaxLogit~\cite{chan2021segmentmeifyoucan} & 36.32 & 91.04 & 82.68 \\
MSP~\cite{hendrycks2018baseline} & 69.85 & 78.24 & 67.93 \\
ODIN~\cite{liang2020enhancing} & 36.32 & 91.04 & 82.68 \\
Mahalanobis~\cite{lee2018simple} & 78.02 & 70.93 & 59.84 \\
LOF~\cite{10.1145/335191.335388} & 76.71 & 67.54 & 55.35 \\
Isolation Forest~\cite{4781136} & 98.64 & 41.94 & 33.50 \\
JointEnergy~\cite{wang2021multilabel} & 31.96 & 92.32 & 86.87 \\ 
\midrule
\textbf{SNoJoE}(ours) & \textbf{28.49} & \textbf{93.48} & \textbf{88.11} \\
\bottomrule
\end{tabular}
\end{subtable}

\begin{subtable}[b]{\textwidth}
\caption{$\mathcal{D}_{out}$ = Texture\cite{cimpoi2013describing}}
\label{tab:texture-ood}
\centering
\setlength{\tabcolsep}{8pt} 
\begin{tabular}{lccc}
\toprule
\textbf{OOD Score} & \textbf{FPR95}$\downarrow$ & \textbf{AUROC}$\uparrow$ & \textbf{AUPR}$\uparrow$ \\ 
\midrule
MaxLogit~\cite{chan2021segmentmeifyoucan} & 12.36 & 96.22 & 96.97 \\
MSP~\cite{hendrycks2018baseline} & 41.81 & 89.76 & 93.00 \\
ODIN~\cite{liang2020enhancing} & 12.36 & 96.22 & 96.97 \\
Mahalanobis~\cite{lee2018simple} & 19.17 & 96.23 & 97.90 \\
LOF~\cite{10.1145/335191.335388} & 89.49 & 60.37 & 76.70 \\
Isolation Forest~\cite{4781136} & 99.59 & 20.89 & 50.11 \\
JointEnergy~\cite{wang2021multilabel} & 10.87 & 96.78 & 97.87 \\ 
\midrule
\textbf{SNoJoE}(ours) & $\textbf{5.02}^\dagger$ & $\textbf{98.48}^\dagger$ & $\textbf{99.00}^\dagger$ \\
\bottomrule
\end{tabular}
\end{subtable}
\end{table}

In Table~\ref{tab:image-ood}, we compare our approach with leading OOD detection methods from the literature, showcasing SNoJoE as the new \textit{state-of-the-art} benchmark. 
Our experimental design carefully selects methods based on pre-trained models to maintain fair comparison standards. Following the guidelines set forth in ~\cite{wang2021multilabel}, we evaluated all metrics using the ImageNet dataset for OOD detection.

Additionally, as detailed in Section~\ref{setup}, we conducted further evaluations using the Texture dataset for OOD detection, with results presented in Table~\ref{tab:texture-ood}. 
Noteworthy, baseline methods like MaxLogit~\cite{chan2021segmentmeifyoucan}, Maximum Softmax Probability (MSP)~\cite{hendrycks2018baseline}, ODIN~\cite{liang2020enhancing}, and Mahalanobis~\cite{lee2018simple} utilize statistics from the highest values across labels to calculate OOD scores. 
The Local Outlier Factor (LOF)~\cite{10.1145/335191.335388} uses K-nearest neighbors (KNN) to assess local densities, identifying OOD samples through their relatively lower densities compared to neighbors. 
The Isolation Forest method~\cite{4781136}, a tree-based strategy, identifies anomalies by the path lengths from root to terminal nodes. 
JointEnergy~\cite{wang2021multilabel} is an energy-based approach that detects OOD instances by evaluating the joint uncertainty among labels.

When conducting OOD detection on different datasets, SNoJoE outperforms several baseline methods across three evaluation metrics. 
% Particularly, when using a subset of ImageNet-22K as the OOD dataset, SNoJoE reduces the FPR95 by 7.83\% compared to MaxLogits, which relies on the maximum value statistics of labels. Furthermore, when using Texture as the OOD dataset, SNoJoE achieves a 7.34\% reduction in FPR95 compared to MaxLogits. 
% Compared to JointEnergy, which performs OOD detection by utilizing label-wise joint energy, SNoJoE achieves a 3.47\% reduction of FPR95 on the subset of ImageNet-22K and a 5.85\% reduction in FPR95 on the Texture dataset.
Compared to JointEnergy, which performs OOD detection by utilizing label-wise joint energy, SNoJoE achieves a 11\% relative reduction of FPR95 on the subset of ImageNet-22K and a 54\% rrelative eduction of FPR95 on the Texture dataset.
% 1 - 28.49/31.96 = 0.1086
% 1 -  5.02/10.87 = 0.5382

\subsection{Ablation Studies}

In this section, we delve into a series of ablation experiments to further affirm that neural networks, when subjected to spectral normalization, exhibit a highly regularized feature space. This regularization, in turn, empowers them to identify generalizable features within the data more effectively, thereby enhancing their capability to accurately distinguish out-of-distribution (OOD) data.
The observed performance improvement of SNoJoE over JointEnergy, as highlighted in Tables~\ref{tab:image-ood} and \ref{tab:texture-ood}, underscores that spectral normalization plays a pivotal role in enabling the extraction of more generalizable features from image input space vectors. This enhancement bolsters the model's proficiency in recognizing OOD samples with greater effectiveness.

In conducting our ablation studies, we persist in utilizing JointEnergy~\cite{wang2021multilabel} as the benchmark for comparison against our method, SNoJoE. This choice is motivated by the findings presented in Section~\ref{results}, where JointEnergy emerged as the most proficient among competing methods, excluding ours. It's noteworthy that both JointEnergy and SNoJoE capitalize on the joint uncertainty between labels to facilitate OOD detection. For the training configurations and parameters, we adhere to the specifications outlined in Section~\ref{setup}.
% and use SNoJoE without spectral normalization ($\mathrm{SNoJoE}^{\dagger}$) as another baseline to validate the performance of spectral normalization in multi-label OOD detection.

% From Tables~\ref{tab:abl-img} and \ref{tab:abl-txt}, it can be observed that SNoJoE achieves a lower FPR95 than the other two baselines on two different OOD datasets, with an average reduction of 2.03\% compared to $\mathrm{SNoJoE}^{\dagger}$. 
% Specifically, $\mathrm{SNoJoE}^{\dagger}$ slightly outperforms JointEnergy in OOD detection on ImageNet-22K. It might be due to the inherent residual connections in the pre-trained ResNet architecture, which allows it to maintain good performance even without spectral normalization. 
% In a word, SNoJoE outperforms JointEnergy regardless of whether spectral normalization is applied, and the performance of the model is further improved with spectral normalization. 

Additionally, we explored the application of spectral normalization across various layers of the network structure to gauge its influence on multi-label OOD detection tasks. 
Our experimental findings, detailed in Tables~\ref{tab:abl_sn_img} and \ref{tab:abl_sn_dtd}, involved implementing spectral normalization at different levels within the ResNet framework \cite{he2016identity} to assess its effect on OOD detection. 
The results suggest that indiscriminate use of spectral normalization could, in some cases, impair the model's ability to perform multi-label OOD detection effectively. 
Specifically, when spectral normalization is limited to the first seven layers of the network (refer to the second row of Tables~\ref{tab:abl_sn_img} and \ref{tab:abl_sn_dtd}), the model's efficacy may decline compared to a non-normalized version. 
This deterioration in performance might stem from the application of spectral normalization solely to the network's more superficial layers. 
Given that these initial layers process simpler representations, imposing stringent constraints on them could diminish the network's capacity for expressive representation, thereby undermining its performance. 
Conversely, extending spectral normalization to the model's deeper layers (as illustrated in the last two rows of Tables~\ref{tab:abl_sn_img} and \ref{tab:abl_sn_dtd}) appears to enhance the model's proficiency in learning and capturing intricate input vector features. 
This improvement is likely due to the advanced abstraction abilities of the deeper layers. 
However, this finding should not be misconstrued to suggest that greater application of spectral normalization invariably results in superior performance, as it also increases computational demands. 
Furthermore, the practicality of applying spectral normalization to certain layers (such as those involved in average pooling to decrease spatial dimensions of feature maps) remains questionable, given the negligible benefits it may offer.

\begin{table}[!tb]
\caption{Ablation study on the impact of the numbers of layers applied spectral normalization using ImageNet-22K (20 classes) and Texture~\cite{cimpoi2013describing} as OOD datasets. \textbf{$\#$layers} are numbers of layers applied spectral normalization.}
\begin{subtable}[b]{\textwidth}
    \caption{$\mathcal{D}_{out}$: ImageNet-22K (20 classes)}
    \label{tab:abl_sn_img}
    \centering
    \setlength{\tabcolsep}{8pt} 
    \begin{tabular}{cccc}
    \toprule
    \textbf{$\#$layers} & \textbf{FPR95}$\downarrow$ & \textbf{AUROC}$\uparrow$ & \textbf{AUPR}$\uparrow$ \\ 
    \midrule
    0 & 31.37 & 93.37 & \textbf{89.29} \\ 
    7 & 48.19 & 89.59 & 84.19 \\ 
    8 & 30.85 & 92.98 & 87.24 \\ 
    9 & \textbf{28.49} & \textbf{93.48} & 88.11 \\ 
    \bottomrule
    \end{tabular}
\end{subtable}

\begin{subtable}[b]{\textwidth}
    \caption{$\mathcal{D}_{out}$: Texture\cite{cimpoi2013describing}}
    \label{tab:abl_sn_dtd}
    \centering
    \setlength{\tabcolsep}{8pt} 
    \begin{tabular}{cccc}
    \toprule
    \textbf{$\#$layers} & \textbf{FPR95}$\downarrow$ & \textbf{AUROC}$\uparrow$ & \textbf{AUPR}$\uparrow$ \\ 
    \midrule
    0 & 6.21 & 97.87 & 98.58 \\ 
    7 & 6.72 & 98.20 & 98.94 \\ 
    8 & \textbf{4.91} & \textbf{98.49} & 98.94 \\ 
    9 & 5.02 & 98.48 & \textbf{99.00} \\ 
    \bottomrule
    \end{tabular}    
\end{subtable}
\end{table}

In summary, our experiments reveal that indiscriminate use of spectral normalization across the network does not invariably enhance the model's performance and might even impair it. 
Nevertheless, if spectral normalization is judiciously applied to enable the network to more effectively learn complex and generalizable features from the input vectors, the model's performance surpasses that of models without spectral normalization. 
The performance discrepancy can reach as high as 2.88 and 1.30 in FPR95, with the OOD dataset being ImageNet-22K and Texture, respectively.

\section{Conclusion}

% In this work, we propose a novel OOD detection method, Spectral Normalized Joint Energy (SNoJoE), based on energy scores in a multi-label classification setting. Our study shows that, before utilizing energy scores to aggregate information across multiple labels, applying spectral normalization to the first several layers of the pre-trained model's network architecture can maintain model robustness, enhance model generalization, and better separate out-of-distribution and in-distribution inputs. Compared to competitive baseline methods, SNoJoE achieves superior OOD detection performance, defining the new \textit{state of the art} in this research area, without incurring a lot more computational cost. 
% We hope that our work will stimulate more attention to multi-label OOD detection tasks and extend this line of research into broader fields.
% Additionally, this method receives ample mathematical explanation from the perspective of joint likelihood, establishing itself as a reliable uncertainty estimation method. 
% Moreover, SNoJoE provides a straightforward approach to OOD detection, without the need for \textit{state-of-the-art} models or high computational costs. 

In this study, we introduce a cutting-edge method for OOD detection named Spectral Normalized Joint Energy (SNoJoE) in the context of multi-label classification. 

Our findings reveal that spectral normalization applied to the initial layers of a pre-trained model's network significantly enhances model robustness, improves generalization capabilities, and more effectively distinguishes between in-distribution and out-of-distribution inputs.
% prior to the utilization of energy scores for aggregating label information 
When compared to leading baseline approaches, SNoJoE sets a new benchmark for OOD detection, establishing itself as the new \textit{state of the art} in this domain, while not substantially increasing computational demands. 

We anticipate that our contribution will spark further exploration into multi-label OOD detection and encourage the expansion of this research area into wider applications.

\begin{credits}
\subsubsection{\ackname}
This work was supported by National Key R\&D Program of China (No. 2021YFC3340700), NSFC grant (No. 62136002), Ministry of Education Research Joint Fund Project (8091B042239), Shanghai Knowledge Service Platform Project (No. ZF1213), and Shanghai Trusted Industry Internet Software Collaborative Innovation Center.
\end{credits}

% \clearpage

\bibliographystyle{splncs04}
\bibliography{custom}

\end{document}